# A Novel Clustering Algorithm Based on a Modified Model of Random Walk

Qiang Li, Yan He, Jing-ping Jiang

College of Electrical Engineering, Zhejiang University,
Hang Zhou, Zhejiang, 310027, China

May 28, 2018

**Abstract**

We introduce a modified model of random walk, and then develop two novel clustering algorithms based on it. In the algorithms, each data point in a dataset is considered as a particle which can move at random in space according to the preset rules in the modified model. Further, this data point may be also viewed as a local control subsystem, in which the controller adjusts its transition probability vector in terms of the feedbacks of all data points, and then its transition direction is identified by an event-generating function. Finally, the positions of all data points are updated. As they move in space, data points collect gradually and some separating parts emerge among them automatically. As a consequence, data points that belong to the same class are located at a same position, whereas those that belong to different classes are away from one another. Moreover, the experimental results have demonstrated that data points in the test datasets are clustered reasonably and efficiently, and the comparison with other algorithms also provides an indication of the effectiveness of the proposed algorithms.

## 1 Introduction

Data clustering is a widely investigated problem in Pattern Recognition. For the past forty years, a lot of excellent algorithms for clustering have been presented from those that put the emphasis on cluster centers and boundaries, say, $K$-means [1], support vector clustering (SVC) [2], to current particle swarm optimization (PSO) based [3], ant-based [4], and flocking-based [5] algorithms for clustering. Observing the history of clustering algorithms, we can notice that a significant change has been made, which may be considered as two stages. First, with fixed data points, we utilized various functions to find complex curve planes in order to cluster or classify data points; second, till the past few years, some pioneers thought about that why not those data points could move in themselves, just like agents or whatever, and collect together automatically. Therefore, following their ideas, they create a few exciting algorithms [3, 4, 5], in which data points moves in a whole space according to certain simple local rules preset in advance.

In addition, a random walk is a special class of stochastic processes [6], which can be simply described in this way. Assume a particle walks on a straight line who either takes one step to the right with probability $p$ or one step to the left with probability $q = 1 - p$ at a time. As a result, the position sequence that is produced by the particle moving is defined as a random walk on a line. Likewise, given a graph and a particle located at one of its vertexes, this particle visits one of its neighbor vertexes at random, and then from this vertex it selects next vertex randomly again. Finally, the sequence of vertexes visited by the particle is defined as a random walk on a graph [7]. Besides, random walks in higher dimensional space and their variations have also been studied, which bring more complex behavior.

In this paper, we propose a modified model of random walk, and develop two clustering algorithms based on it. Furthermore, in our algorithms, data points in a dataset are considered as particles that can walk in space at random. Further, each data point can also be viewed as a local control subsystem, whose controller controls its walking behavior. After taking a step to one of its neighbors, its position will be updated. As data points moves in space at random according to the rules of the modified model, they gather together gradually, and finally form some clusters automatically. The remainder of this paper is organized as follows: Section 2 reviews some related work about random walks briefly, and explains our motivation. In Section 3 the modified model of random walk is introduced specifically. In Section 4 the convergence of modified model of random walk is discussed. In Section 5 firstly two clustering algorithms based on the model are elaborated, and then they are analyzed in detail. Next, the effects of some important parameters are discussed. In section 6 experimental results of algorithms are demonstrated. Finally, the conclusion is given in Section 7.

# 2 Related work

Our work about dynamical clustering is inspired by Cui et al. [5], who present a flocking based algorithm for document clustering. In their algorithm, each document vector is mapped as a boid into a two-dimensional virtual space firstly. By means of four rules, boids with similar features move, collect together automatically and establish a flock, while flocks with different features keep away from one another.

In the last ten years, the methods related to random walk have been wildly applied in all kinds of fields, such as computer science [8, 9, 10, 11], physics [12], and biology [13]. Especially, in the algorithmic theory, some approximate algo-rithms based on random walk have been presented to solve NP-hard problems.

On the other hand, although random walks on graphs [7] as a theory have been investigated by mathematicians for some years, the idea does not be applied to the domain of pattern recognition by some researchers until recent years. For instance, Luh Yen et al. [8] gave a random walk based distance measure that is applied in the $k$-means algorithm as a new distance measure. David Harel et al. [9] studied an algorithm based on deterministic exploration of random walks on a weighted graph. The similarities between data points were computed based on the $k$-th power of transition probability matrix. Consequently, edges with similarities zero or approaching zero were separated by *separating operators*. Thus, each non-connected subgraph represented a cluster of spatial data. Later, Günes Erkan [10] extended the model similar to David Harel's to the directed graph case, and introduced a language model-based document clustering algorithm as an application of this model.

In their algorithms, however, the vertexes of the graph (data points) are located at some positions fixedly, so the shape of graph which represents the connections among data points is almost unchanged. Unlike their methods, data points in our algorithms are regarded as not only vertexes of a graph but also particles which can move in the whole space according to the rules of the modified model, so that the shape of the graph that they construct is changed over time. As data points walk constantly at random in space, as a consequence, the clusters are established automatically in the process.

# 3 Modified model of random walk

Assume a set $\boldsymbol{X}$ with $N$ particles, $\boldsymbol{X} = \{\boldsymbol{X}_1, \boldsymbol{X}_2, \cdots, \boldsymbol{X}_N\}$, in which $\boldsymbol{X}_i$ is the position of a particle located in an $m$-dimensional metric space. From the point of view of control theory, the system composed of $N$ particles which walk randomly in space may be described by the below block diagram Fig. 1.

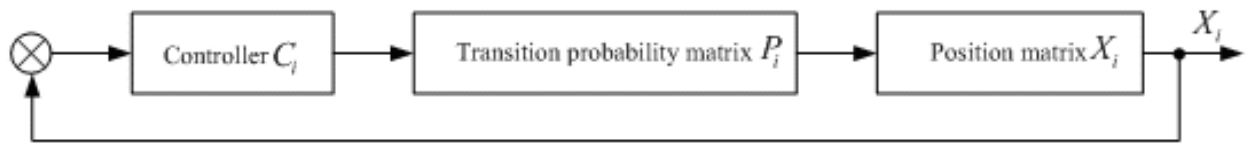

Figure 1: Block diagram of the system.

As is shown in Fig. 1, the controlled object is the Transition Probability matrix $P$, and the outputs of the system are the new positions of all particles in the system. The controller $C$ adjusts the entries in the Transition Probability matrix $P$ according to the current positions of $N$ particles, and then decides the transition directions and transition distances of $N$ particles at the next moment. Finally, the positions of $N$ particles are updated synchronously. So, the equation of motion of the whole system and the output equation are written

as follows:

$$
\begin{cases}
P(t+1)_{n\times n} = [P_i(t+1)_{1\times n}]_{i=1,2,\cdots,N} = b_{n\times n} \oslash (b_{n\times n} \times (1)_{n\times n}) \\
b_{n\times n} = (L(t+1)_{n\times n} \boxtimes (K(t+1)_{n\times 1} \times (1)_{1\times n}) \boxtimes (K(0)_{n\times 1} \times (1)_{1\times n})) \\
\oslash (L(t+1)_{n\times n} \times (K(t+1)_{n\times 1} \boxtimes K(0)_{n\times 1}) \times (1)_{1\times n}) \\
\oslash (D(t+1)_{n\times n} \boxtimes D(0)_{n\times n}) \\
\boldsymbol{X}(t+1)_{n\times m} = [\boldsymbol{X}_i(t+1)_{1\times m}]_{i=1,2,\cdots,N} = \boldsymbol{X}(t)_{n\times m} + (((Eve(t+1)_{n\times n} \\
\boxtimes P(t+1)_{n\times n} \boxtimes D(t+1)_{n\times n}) \times (1)_{n\times 1}) \times (1)_{1\times m}) \\
\boxtimes (\boldsymbol{X}(t)_{n\times m} - Eve(t+1)_{n\times n} \times \boldsymbol{X}(t)_{n\times m})
\end{cases}
\quad (1)
$$

where the matrix $D(t+1)_{n\times n}$ and $L(t+1)_{n\times n}$ represent a distance matrix and an adjacent matrix of $N$ particles respectively; $Eve(t+1)_{n\times n}$ is an event matrix, which indicates the transition directions of particles in the system; $K(t+1)_{n\times 1}$ denotes the degree vector, each of which describes the number of particles within the neighborhood of a particle $\boldsymbol{X}_i$. Besides, we have also defined two new matrix operations which are expressed by Eq. 2: (a) the multiplication of corresponding entries of two matrixes represented by symbol '$\boxtimes$', and (b) the division of corresponding entries of two matrixes represented by symbol '$\oslash$'.

$$
\begin{aligned}
A_{n\times n} \boxtimes B_{n\times n} &= \begin{pmatrix} a_{11}\times b_{11} & \cdots & a_{1n}\times b_{1n} \\ \vdots & \vdots & \vdots \\ a_{11}\times b_{n1} & \cdots & a_{nn}\times b_{nn} \end{pmatrix}, \\
A_{n\times n} \oslash B_{n\times n} &= \begin{pmatrix} a_{11}/b_{11} & \cdots & a_{1n}/b_{1n} \\ \vdots & \vdots & \vdots \\ a_{11}/b_{n1} & \cdots & a_{nn}/b_{nn} \end{pmatrix}.
\end{aligned}
\quad (2)
$$

Further, if each particle $\boldsymbol{X}_i$ is viewed as a local control subsystem, the whole system mentioned previously can be redrawn as a new distributed system composed of $N$ local control subsystems, shown in Fig. 2, where the output of each subsystem is fed back to all other subsystems besides itself.

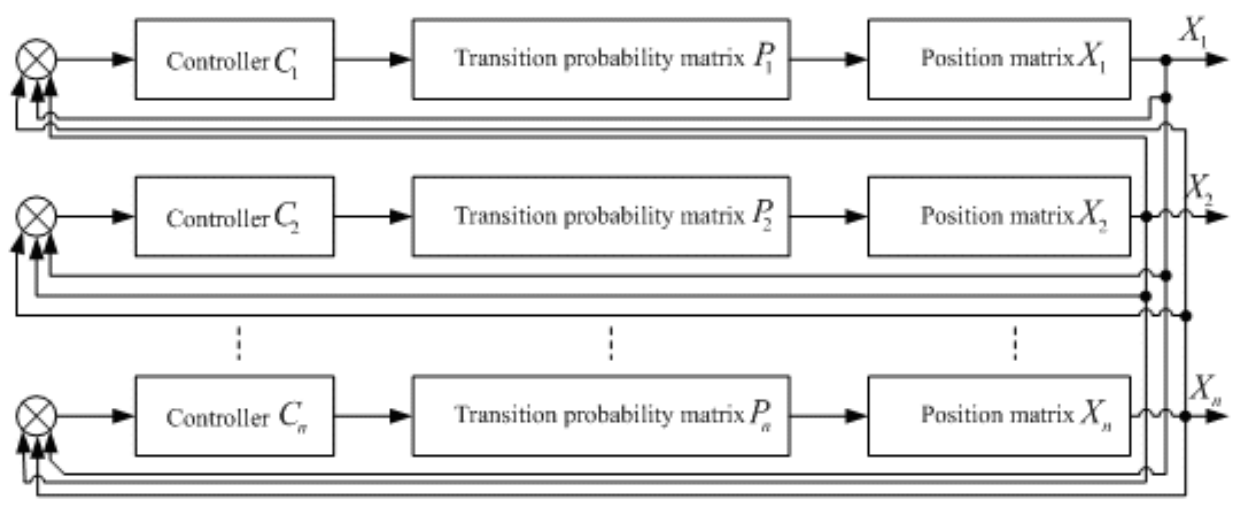

Figure 2: Block diagram of $N$ local control subsystems.

As for a local control subsystem about a particle $\boldsymbol{X}_i$, the controller $C_i$ gets its own as well as other particles' current positions by receiving feedbacks of all subsystems, and then computes the distances between it and all other particles to form a distance vector $D_i(t+1)_{1\times n}$ by means of the selected distance function $d : \boldsymbol{X} \times \boldsymbol{X} \longrightarrow \mathbb{R}$ which satisfies the closer the two particles are, the smaller the output of the function is.

$$D_i(t+1)_{1\times n} = \Big[d\Big(\boldsymbol{X}_i(t), \boldsymbol{X}_j(t)\Big)\Big]_{1\times n}, j = 1, 2, \cdots, N \tag{3}$$

Based on the distance vector $D_i(t+1)_{1\times n}$ and setting an interaction range $R$ for each particle, the adjacent vector $L_i(t+1)_{1\times n}$ of a particle $\boldsymbol{X}_i$ may be established by Eq. 4, which indicates how many particles are within its $R$ neighborhood.

$$\begin{aligned} L_i(t+1)_{1\times n} &= \Big\{l_{ij}(t+1), j = 1, 2, \cdots, N\Big\} \\ l_{ij}(t+1) &= \begin{cases} 1 & \text{if } d\Big(\boldsymbol{X}_i(t), \boldsymbol{X}_j(t)\Big) \le R \\ 0 & \text{otherwise} \end{cases} \end{aligned} \tag{4}$$

Furthermore, the neighbor set $\Gamma_i(t+1)$ and the degree of the particle $\boldsymbol{X}_i$ are produced according to the adjacent vector $L_i(t+1)_{1\times n}$:

$$\begin{aligned} \Gamma_i(t+1) &= \Big\{j| \text{ if } l_{ij}(t+1) = 1\Big\} \\ K_i(t+1) &= \textstyle\sum_{j\in\Gamma_i(t+1)} l_{ij}(t+1) = \Big|\Gamma_i(t+1)\Big| \end{aligned} \tag{5}$$

Here, the symbol $|\cdot|$ represents the cardinality of a set. Thus, the distance matrix $D(t+1)_{n\times n}$, adjacent matrix $L(t+1)_{n\times n}$ and the degree vector $K(t+1)_{n\times 1}$ of the whole system take forms as following:

$$\begin{cases} D(t+1)_{n\times n} = \Big[D_i(t+1)_{1\times n}\Big] \\ L(t+1)_{n\times n} = \Big[L_i(t+1)_{1\times n}\Big] \quad , i = 1, 2, \cdots, N \\ K(t+1)_{n\times n} = \Big[K_i(t+1)\Big] \end{cases} \tag{6}$$

Before the particle $\boldsymbol{X}_i$ walks, the transition probabilities of which moves to all neighbors in its neighborhood need to be computed firstly as below:

$$\begin{aligned} P_i(t+1)_{1\times n} &= \Big\{p_{ij}(t+1), j = 1, 2, \cdots, N\Big\} \\ p_{ij}(t+1) &= \begin{cases} \frac{a_{ij}(t+1)}{\sum_{j\in\Gamma_i(t+1)} a_{ij}(t+1)} & \text{if } j \in \Gamma_i(t+1) \\ 0 & \text{otherwise} \end{cases} \\ a_{ij}(t+1) &= \frac{\Big(K_j(t+1)/\sum_{j\in\Gamma_i(t+1)} K_j(t+1)\Big) \times \Big(K_j(0)/\sum_{j\in\Gamma_i(0)} K_j(0)\Big)}{\Big(d\Big(\boldsymbol{X}_i(t), \boldsymbol{X}_j(t)\Big)\Big) \times \Big(d\Big(\boldsymbol{X}_i(0), \boldsymbol{X}_j(0)\Big)\Big)} \end{aligned} \tag{7}$$

where $K_j(0)$ and $d(\boldsymbol{X}_i(0), \boldsymbol{X}_j(0))$ are the initial degree of the particle $\boldsymbol{X}_i$ and the initial distance between two particles $\boldsymbol{X}_i$ and $\boldsymbol{X}_j$ respectively.

For a particle in the system, it is allowed to choose only one of neighbors in its neighborhood as the transition direction at a time, and then takes a step toward this neighbor. Eventually, which neighbor is selected as the transition direction by the particle $\boldsymbol{X}_i$? This depends on an event-generating function $G_i(\cdot, \cdot)$, which is a function of transition probability vector $P_i(t+1)_{1\times n}$ and distance vector $D_i(t+1)_{1\times n}$ of the particle $\boldsymbol{X}_i$. Before using the event-generating function, a set of events of the particle $\boldsymbol{X}_i$, $Eve_i(t+1)_{1\times n} = \{eve_{ij}, j = 1, 2, \cdots, N\}$ need to be built at first, whose each element, an event $eve_{ik}, k \in \Gamma_i(t+1)$, corresponds to a neighbor in its neighborhood. The event generated by the event-generating function indicates an event in the event set $Eve_i(t+1)$ takes place.

$$Eve_i(t+1)_{1\times n} = \Big\{eve_{ij}(t+1), j = 1, 2, \cdots, N\Big\}$$
$$\begin{cases} eve_{ik}(t+1) = 1 & \text{if } G_i\Big(P_i(t+1), D_i(t+1)\Big) = k, \\ & \quad k \in \Gamma_i(t+1) \\ eve_{ij}(t+1) = 0 & j \in \Gamma_i(t+1)\backslash k \end{cases} \tag{8}$$

It is worth noting that the event-generating function $G_i(\cdot, \cdot)$ generates only one event at a time, that is, only one corresponding event in the event set $Eve_i(t+1)_{1\times n}$ occurs exactly. As such, the particle $\boldsymbol{X}_i$ will take a step toward the neighbor $\boldsymbol{X}_k$, and the walking length $\omega_i(t+1)$ is proportional to the transition probability $p_{ik}(t+1)$, as is expressed below:

$$\begin{aligned} \omega_i(t+1) &= p_{ik}(t+1) \times d\Big(\boldsymbol{X}_i(t), \boldsymbol{X}_k(t)\Big) \\ &= \Big(P_i(t+1)_{1\times n} \times \Big(Eve_i(t+1)_{1\times n}\Big)^T\Big) \\ &\times \Big(D_i(t+1)_{1\times n} \times \Big(Eve_i(t+1)_{1\times n}\Big)^T\Big) \end{aligned} \tag{9}$$

After the particle $\boldsymbol{X}_i$ walks, its position will be updated by means of the following equation:

$$\begin{aligned} \boldsymbol{X}_i(t+1)_{1\times m} &= \boldsymbol{X}_i(t)_{1\times m} + (Eve_i(t+1)_{1\times n} \times \boldsymbol{X}(t)_{n\times m} - \boldsymbol{X}_i(t)_{1\times m}) \\ &\quad \times \omega_i(t+1)/d\Big(\boldsymbol{X}_i(t), \boldsymbol{X}_k(t)\Big) \\ &= \boldsymbol{X}_i(t)_{1\times m} + (Eve_i(t+1)_{1\times n} \times \boldsymbol{X}(t)_{n\times m} - \boldsymbol{X}_i(t)_{1\times m}) \times p_{ik}(t+1) \end{aligned} \tag{10}$$

When all particles in the system have walked, their positions will be recomputed synchronously, as means that an iteration of the modified model is completed.

# 4 Convergence of modified model

In this section, we attempt to discuss the convergence of the modified model of random walk. As for the system with $N$ particles walking randomly governed by the modified model, they will walk, collect and form several separating clusters eventually, if the convergence of the model holds. At first, we introduce Theorem 1 [14], which specifies the absorbing probability of a particle $A$ at the origin, when it walks at random on a line.

**Theorem 1** *Particle $A$ is a particle walking randomly on a line. $A(n)$ represents the position of Particle $A$ at moment $n$, $\{A(n), n = 0, 1, 2, \cdots\}$, and $A(0)$ is the initial position of Particle $A$, $A(0) = l, l = 0, 1, 2, \cdots$. Its transition probabilities are defined as following,*

$$P_{l,i} = \begin{cases} p_l > 0 & \text{if } i = l+1 \\ r_l > 0 & \text{if } i = l \\ q_l > 0 & \text{if } i = l-1 \\ 0 & \text{otherwise} \end{cases} \tag{11}$$

*Position 0, the origin, is an absorbing status, that is, if the current position of Particle $A$ lies at the origin, then the walk is stopped. Further, if the transition probabilities are constants, then the probability of Particle $A$ absorbed at Position 0 is:*

$$f_{l,0} = \sum_{i=0}^{\infty} f_{i,0} = \begin{cases} (q/p)^l & \text{if } p > q \\ 1 & \text{if } p \le q \end{cases} \tag{12}$$
$$f_{0,0} = 1$$

**Proof 1** *see reference [14].*

According to Theorem 1, if the transition probability of Particle $A$ walking to the origin is larger than that of moving conversely, or the probabilities of moving in two opposite directions are equal, then Particle $A$ will be absorbed at the origin with probability one. On the other hand, if the transition probability of moving away from the origin is larger than the other transition probability, the absorbing probability drops to $(q/p)^l$. Further, let's regard two extreme cases: $(a)$ when the transition probability of moving away from the origin is much larger than the other transition probability, $p \gg q$; or $(b)$ when the initial position of Particle $A$ approaches infinite and $p > q$, the limit of absorbing probability of Particle $A$ is zero, as is expressed below:

$$(a) f_{l,0} = \lim_{q/p \to 0} \left(\frac{q}{p}\right)^l = 0 \text{ or } (b) f_{l,0} = \lim_{l \to \infty} \left(\frac{q}{p}\right)^l = 0 \tag{13}$$

**Theorem 2** *Particle $A$ and $B$ are two particles walking randomly on a line. $A(n)$ and $B(n)$ represent the positions of Particle $A$ and $B$ at moment $n$ respectively, $\{A(n), n = 0, 1, 2, \cdots\}$, $\{B(n), n = 0, 1, 2, \cdots\}$. $A(0)$ and $B(0)$ are the initial positions of Particle $A$ and $B$, $A(0) = j, B(0) = k, j, k = 0, 1, 2, \cdots$. Their transition probabilities are defined as below,*

$$P^a_{j,i} = \begin{cases} p^a_j > 0 & \text{if } i = j+1 \\ q^a_j > 0 & \text{if } i = j-1 \\ 0 & \text{otherwise} \end{cases}, P^b_{k,i} = \begin{cases} p^b_k > 0 & \text{if } i = k+1 \\ q^b_k > 0 & \text{if } i = k-1 \\ 0 & \text{otherwise} \end{cases} \tag{14}$$

*Hence, the encounter probability of Particle A and B is:*

$$f_{l,0}^z = \begin{cases} (p_j^a q_k^b / q_j^a p_k^b) & \text{if } q_j^a p_k^b > p_j^a q_k^b \\ 1 & \text{if } q_j^a p_k^b \le p_j^a q_k^b \end{cases} \tag{15}$$
$$f_{0,0}^z = 1$$

**Proof 2** *when Particle A encounters Particle B at certain position on a line, the distance between them must be zero. Therefore, at first we define a variable* $Z(n)$*, which represents the distance between Particle A and B,* $\{Z(n) = A(n) - B(n), n = 0, 1, 2, \cdots\}$*. At the beginning, the distance between Particle A and B is* $Z(0) = k - j = l, l = 2, 4, \cdots, 2n, \cdots$*, whose value must be an even. This is because at each step both Particle A and B move a unit, if the initial distance is an odd, then no matter how particles walk and how long the time is taken, they won't encounter on this line, i.e.,* $Z(i) \neq 0, i = 1, 2, \cdots$*. To solve this problem, we introduce a new variable* $\eta = 1$*. If the distance between these two particles is less than or equal to* $\eta$*, we deem they encounter at that time. Thus, the initial distance* $l$ *may be an even or an odd.*

*If considering the sequence* $\{Z(n), n = 0, 1, 2, \cdots\}$ *as a position sequence of Particle Z, then transition probabilities of Particle Z are associated with the transition probabilities of Particle A and B, which take the form as below,*

$$P_{l,i}^z = \begin{cases} p_j^a q_k^b & \text{if } i = l - 2 \\ p_j^a p_k^b + q_j^a q_k^b & \text{if } i = l \\ q_j^a p_k^b & \text{if } i = l + 2 \end{cases} \tag{16}$$

*Therefore, the sequence* $Z(n)$ *is a random walk on a line. As such, the encounter probability of Particle A and B is equal to the probability of Particle Z absorbed at the origin. So, according to Theorem 1, the absorbing probability of Particle Z, i.e., the encounter probability of Particle A and B, is:*

$$f_{l,0}^z = \begin{cases} (p_j^a q_k^b / q_j^a p_k^b) & \text{if } q_j^a p_k^b > p_j^a q_k^b \\ 1 & \text{if } q_j^a p_k^b \le p_j^a q_k^b \end{cases}$$
$$f_{0,0}^z = 1$$

*The Theorem is proved.*

According to our modified model, a particle $\boldsymbol{X}_i$ only can take one step $\omega_i$ to one of its neighbors in its neighbor set $\Gamma_i(t+1)$ at a time. If the particle $\boldsymbol{X}_i$ has taken a step to a particle $\boldsymbol{X}_j$ which is one of its neighbors, and the particle $\boldsymbol{X}_j$ walks to $\boldsymbol{X}_i$ as well or stays still, then the distance between them decreases. Applying Eq. 7 to computing the transition probabilities of the particle $\boldsymbol{X}_i$ and $\boldsymbol{X}_j$ respectively, we can see that their transition probabilities moving to each other, $p_{ij}$ and $p_{ji}$, both increase, if their degrees $K_i$ and $K_j$ change slowly enough. During iterations of the model, if this transition process always occurs, according to Theorem 2, the particle $\boldsymbol{X}_i$ and $\boldsymbol{X}_j$ will encounter with probability one, because the inequality, $p_{ik}p_{ih} \leq p_{ij}p_{ji}, k \in \Gamma_i(t+1), h \in \Gamma_j(t+1)$, always holds.

On the other hand, if the particle $\boldsymbol{X}_i$ walks to $\boldsymbol{X}_j$, the distances between the particle $\boldsymbol{X}_i$ and its other neighbors will rise at the same time. As such, the transition probabilities walking to other neighbors will drop too, assuming the degrees still change slowly enough. Simultaneously, if the particle $\boldsymbol{X}_k$ within the neighbor set of the particle $\boldsymbol{X}_i$ is moving away from the particle $\boldsymbol{X}_i$, then the probabilities approaching each other will decrease further according to Eq. 7. If this process is carried to the extreme, the distance between the particles $\boldsymbol{X}_i$ and $\boldsymbol{X}_k$ will increase largely, while their transition probabilities moving to one another will decrease shapely. In this case, the limit of encounter probability is:

$$\begin{aligned}&f^z_{d(\boldsymbol{X}_i(t),\boldsymbol{X}_j(t)),0} = \lim_{d(\boldsymbol{X}_i(t),\boldsymbol{X}_j(t))\to\infty} \left(\frac{p_{ik}p_{ki}}{p_{ij}p_{kh}}\right)^{d(\boldsymbol{X}_i(t),\boldsymbol{X}_j(t))} = 0\\&(\because p_{ij}p_{kh} > p_{ik}p_{ki}), j \in \Gamma_i(t+1), h \in \Gamma_j(t+1)\end{aligned} \tag{17}$$

As analyzed above, for the system with $N$ particles governed by the modified model of random walk, some particles will be close to each other, whereas others will be away from one another. Thanks to the convergence of the modified model, this explains why particles in the particle set $X$ can gather together and establish several separating clusters at last. Although the real processes of motions of particles are far more complex than those simple cases analyzed previously, Theorem 2 still provides a way to analyze and explain the results obtained by the modified model.

# 5 Application to clustering

In the section, at first two clustering algorithms based on the modified model of random walk are constructed, one of which is a deterministic random walk based clustering algorithm (RW1), and the other is a nondeterministic clustering algorithm (RW2). And then two clustering algorithms are analyzed in detail.

## 5.1 Algorithms

Assume an unlabeled dataset $\boldsymbol{X} = \{\boldsymbol{X}_1, \boldsymbol{X}_2, \cdots, \boldsymbol{X}_N\}$, whose each instance is with $m$ features. In the two clustering algorithms based on the modified model, each data point in the dataset is regarded as a movable particle which can walk in the whole space at random and obey the rules given by the modified model of random walk.

After selecting a similarity (or distance) function $d : \boldsymbol{X} \times \boldsymbol{X} \longrightarrow \mathbb{R}$, one can compute the distance matrix $D(t+1)_{n\times n}$, adjacent matrix $L(t+1)_{n\times n}$ and transition probability matrix $P(t+1)_{n\times n}$ step by step according to the modified model. However, if the event-generating function $G_i(\cdot,\cdot)$ is chosen differently, the transition direction and distance of a particle $\boldsymbol{X}_i$ will be diverse. As a consequence, the clustering results may be changed largely depending on different event-generating functions $G_i(\cdot,\cdot)$. Thus, we design two different event-generating functions and then construct two clustering algorithms: (a) a deterministic random walk based clustering algorithm (RW1) and (b) a nondeterministic random walk based clustering algorithm (RW2).

### 5.1.1 Deterministic clustering algorithm (RW1)

The algorithm RW1 using a deterministic event-generating function $G_i^1(\cdot,\cdot)$ has no randomness, i.e., under the same conditions, no matter how many times the algorithm is run, the clustering results obtained are unchanged. The event that is produced by the event-generating function $G_i^1(\cdot,\cdot)$ satisfies the following equation:

$$
\begin{aligned}
& G_i^1(P_i(t+1), D_i(t+1)) = k \\
& = \underset{k \in \Gamma_i(t+1)\setminus\alpha}{argmax} \Big(P_i(t+1) \setminus \{p_{ij}, j \in \alpha\}\Big) \\
& \alpha = \Big\{ d\Big(\boldsymbol{X}_i(t), \boldsymbol{X}_j(t)\Big) \le \theta, j \in \Gamma_i(t+1) \Big\}
\end{aligned}
\tag{18}
$$

where $\theta$ is a collision-avoiding threshold, which indicates the minimal distance between two data points.

The event generated by event-generating function $G_i^1(\cdot,\cdot)$ represents that an event $eve_{ik}$ in event set $Eve_i(t+1)_{1\times n}$ of a data point $\boldsymbol{X}_i$ occurs, namely $eve_{ik} = 1$, which means the neighbor $\boldsymbol{X}_k$ is chosen as the transition direction that is with maximal transition probability and satisfies the inequality $d(\boldsymbol{X}_i(t), \boldsymbol{X}_j(t)) > \theta$. Then, the transition distance is computed by Eq. 9. When corresponding events of all data points are produced, the event matrix of the system is formed, $Eve(t+1)_{n\times n} = [Eve_i(t+1)_{1\times n}]_{i=1,2,\cdots,N}$. Finally, the new position of the data point $\boldsymbol{X}_i$ is updated by means of Eq. 10. After one iteration, the sum of transition distances of all data points is computed, $\sum_{i=1}^{n} \omega_i$. If it is less than a preset threshold $\varepsilon$, all data points stop walking and the algorithm is end.

### 5.1.2 Nondeterministic clustering algorithm (RW2)

The nondeterministic algorithm based on the modified model is characterized by its randomness, which means that wider areas may be explored, and as a result better solutions may be found. In algorithm RW2, the event-generating function with uncertainty is employed to establish the nondeterministic clustering algorithm.

As same as algorithm RW1, the distance matrix $D(t+1)_{n\times n}$, adjacent matrix $L(t+1)_{n\times n}$ and transition probability matrix $P(t+1)_{n\times n}$ need to be computed firstly in terms of the modified model. Next, as for each data point $\boldsymbol{X}_i$, a biased dice with $|\Gamma_i(t+1)|$ faces is applied, whose every face corresponds to a neighbor in its neighbor set. The bias denotes that the probability of every face appearing equals to an nonzero element in the transition probability vector $P_i(t+1)_{1\times n}$. After playing this dice $K_i(t+1)$ times, the results are recorded in a vector $h_i(t+1)$. In practice, the method that we use to simulate the process of playing dice is to divide an interval $[0, 1]$ into $|\Gamma_i(t+1)|$ subintervals, each of which corresponds to a face of the dice, i.e., a neighbor in the neighbor set of the data point $\boldsymbol{X}_i$. The length of each subinterval is equal to a transition probability, $length(g, g+1) = p_{ij}(t+1), g = 0, 1, \cdots, |\Gamma_i(t+1)|, j \in \Gamma_i(t+1)$. Finally, $K_i(t+1)$ random numbers between zero and one are generated according to uniformly distribution, and the number of falling to each subinterval is recorded into the vector $h_i(t+1)$.

As for the event-generating function $G_i^2(\cdot, \cdot)$ applied in algorithm RW2, the event generated by it satisfies the below equation,

$$\begin{aligned} & G_i^2(P_i(t+1), D_i(t+1)) = k \\ & = \underset{k \in \Gamma_i(t+1) \setminus \beta}{argmax} \Big(h_i(t+1) \setminus \{h_j, j \in \beta\}\Big) \\ & \beta = \Big\{ d\Big(\boldsymbol{X}_i(t), \boldsymbol{X}_j(t)\Big) \leq \theta, j \in \Gamma_i(t+1) \Big\} \end{aligned} \tag{19}$$

In other words, the corresponding event $eve_{ik} = 1$ in the event vector $Eve_i(t+1)_{1 \times n}$ takes place, i.e., the neighbor $\boldsymbol{X}_k$ is chosen, where the distance between the data point $\boldsymbol{X}_i$ and $\boldsymbol{X}_k$ is larger than the threshold $\theta$ and the falling number is largest. Next, the data point $\boldsymbol{X}_i$ takes a step $\omega_i$ to the data point $\boldsymbol{X}_k$, and then its position is updated by Eq. 10. Similarly, when the sum of transition distances of all data points is less than a threshold $\varepsilon$, $\sum_{i=1}^{n} \omega_i < \varepsilon$, the algorithm exits.

The steps of Algorithm RW1 and Algorithm RW2 are summarized in Table 1.

Table 1: Steps of clustering algorithm.

| |
|---|
| Select a distance function $d(\cdot, \cdot)$ |
| Initialization: |
| Set interaction range $R$ |
| Compute initial similarity(distance) matrix $D(0)_{n \times n} = [d(\boldsymbol{X}_i(0), \boldsymbol{X}_j(0))]_{i,j=1,2,\cdots,N}$ |
| Compute initial adjacent matrix $L(0)_{n \times n}$ |
| Compute initial degree vector $K(0)_{n \times 1}$ |
| Repeat: |
| Compute current similarity(distance) matrix $D(t+1)_{n \times n}$ using Eq. 3 |
| Produce current adjacent matrix $L(t+1)_{n \times n}$ based on $D(t+1)_{n \times n}$ using Eq. 4 |
| Compute current neighbor set of each data point $\Gamma_i(t)$ and degree vector $K(t+1)_{n \times 1}$ using Eq. 5 |
| For each data points $\boldsymbol{X}_i$ |
| Compute transition probability vector $P_i(t+1)_{1 \times n}$ by means of Eq. 7 |
| RW1: Generate an event using Eq. 18 |
| RW2: Generate an event using Eq. 19 |
| Identify the transition direction of each data point |
| Produce event vector $Eve_i(t+1)_{1 \times n}$ using Eq. 8 |
| Compute transition distance $\omega_i$ using Eq. 9 |
| End For |
| Update positions of all points by means of Eq. 10 |
| Until $\sum_{i=1}^{n} \omega_i < \varepsilon$ |

## 5.2 Analysis of algorithm

In the proposed clustering algorithms, there are two most important things: (a) the computation of transition probability matrix $P_{n\times n}$ and (b) the design of the event-generating function $G_i(\cdot,\cdot)$. Applying different formulations to computing the transition probability matrix or designing different event-generating functions, one can establish various algorithms. As for the problem of data clustering, however, the computation of transition probability matrix, generally speaking, is associated with the similarity (or distance) matrix $D_{n\times n}$. For example, only considering the distances among data points, the formulation for computing the transition probability may be written as:

$$P(t+1)_{n\times n} = [p_{ij}(t+1)]_{i,j=1,2,\cdots,N}$$
$$p_{ij}(t+1) = \begin{cases} \dfrac{1/d\Big(\boldsymbol{X}_i(t),\boldsymbol{X}_j(t)\Big)}{\sum_{j\in\Gamma_i(t+1)} 1/d\Big(\boldsymbol{X}_i(t),\boldsymbol{X}_j(t)\Big)} & \text{if } j\in\Gamma_i(t+1) \\ 1 & \text{otherwise} \end{cases} \tag{20}$$

When all data points begin to walk according to the rules in the modified model, the new positions of data points as compared to their initial positions may be largely changed, in particular for those data points on the boundary of classes. For instance, a boundary point $\boldsymbol{X}_i$ that belongs to a class $y_1$ may appear near a data point $\boldsymbol{X}_j$ that belongs to another class $y_2$, because of choosing a wrong transition direction. As is shown in Fig. 3(a), there are two classes, $y_1$ and $y_2$, and two boundary points, $\boldsymbol{X}_i$ and $\boldsymbol{X}_j$, that belong to two different classes. If the transition probability matrix is computed by means of Eq. 20, the initial transition probabilities of the data point are shown in Fig. 3(b), in which the denominator of a number represents the distance between two data points, and the numerator represents the transition probability. When the data point $\boldsymbol{X}_i$ takes a step to $\boldsymbol{X}_j$, and becomes closer to the point $\boldsymbol{X}_j$, the transition probabilities will increase further when recomputed by Eq. 20, shown in Fig. 3(c). Thus, it is predicted that data points $\boldsymbol{X}_i$ and $\boldsymbol{X}_j$ would encounter at last, and the data point $\boldsymbol{X}_i$ would be clustered wrongly as a point in the class $y_2$.

Analyzing this process carefully, we can find that this is because the computation of transition probabilities only depends on the distances between data points. Once the distance between two data points decreases, at the same time the probabilities attracting each other increase. It is worse that the process of encounter of two data points is accelerated owing to the positive feedback. This phenomenon is also consistent with the fact that is described by Theorem 2. According to Theorem 2, when the product of probabilities of moving to one another is larger than that of moving away from each other, $p_{ik}p_{ih} \le p_{ij}p_{ji}, k\in\Gamma_i(t+1), h\in\Gamma_j(t+1)$, the encounter probability of the data points $\boldsymbol{X}_i$ and $\boldsymbol{X}_j$ will be one, $f^z_{d(\boldsymbol{X}_i(t),\boldsymbol{X}_j(t)),0} = 1$. Hence, we can draw a conclusion that points tend to walk to their neighbors with minimal distances, when choosing Eq. 20 as the formulation of computing transition probability

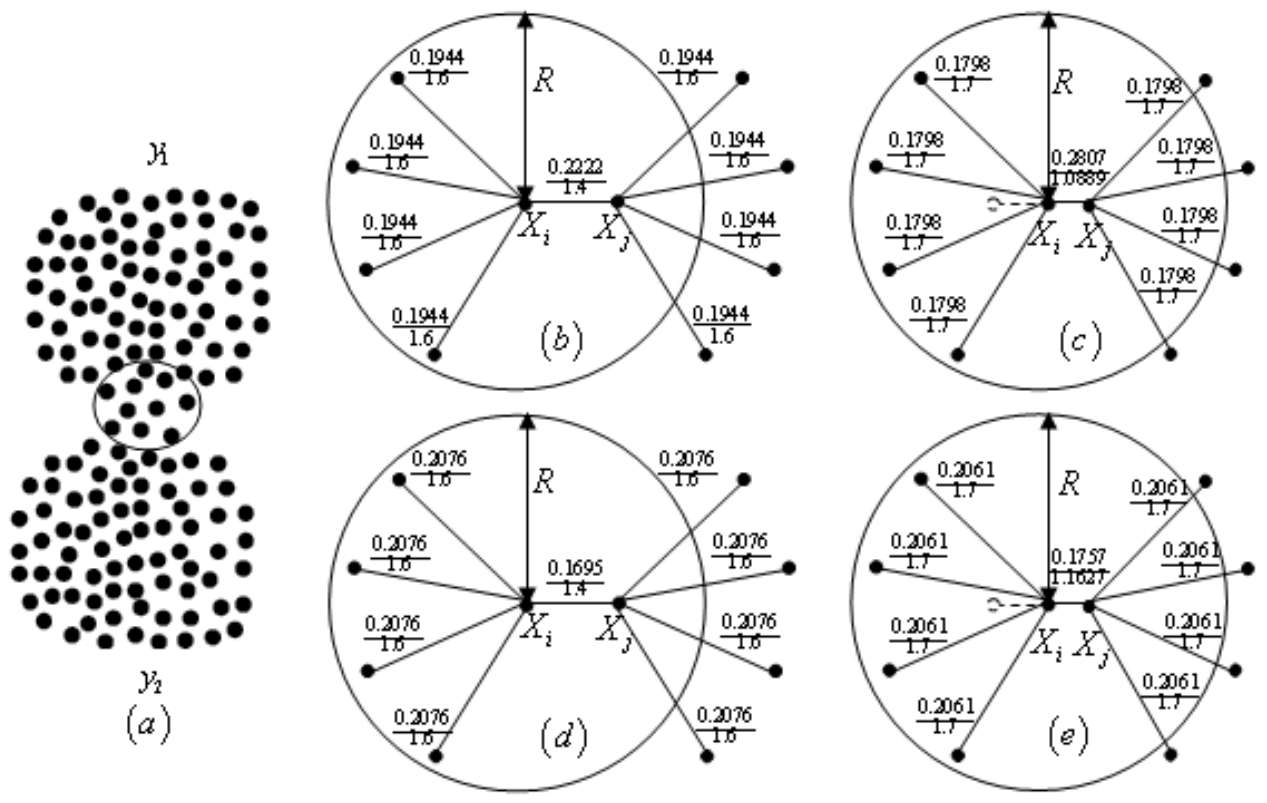

Figure 3: Comparison of two methods for computing transition probabilities.

matrix.

To avoid those problems mentioned above, we introduce a new formulation Eq. 7 to compute the transition probability matrix in the modified model, which is associated with not only the current distance matrix $D(t+1)_{n\times n}$, but also the current degree vector $K(t+1)_{n\times 1}$, the initial distance matrix $D(0)_{n\times n}$ and initial degree vector $K(0)_{n\times 1}$. The degree of a data point $\boldsymbol{X}_i$ describes the number of its neighbors in its neighborhood, and also reflects the distribution of density around the data point $\boldsymbol{X}_i$. As a general rule, a data point with a large degree lies in an area of high density, which indicates this point is perhaps a central point; on the other hand, a point with a small degree may be a boundary point because of low density.

Reanalyzing the above example with the same settings, further we assume the degrees of data points $\boldsymbol{X}_i$ and $\boldsymbol{X}_i$ are $K_i = K_j = 5$, and the degrees of other points are seven. This is logical, since other points are inner points within a class, they could have higher degrees than boundary points $\boldsymbol{X}_i$ and $\boldsymbol{X}_j$. Now, we recalculate the transition probability of the data point $\boldsymbol{X}_i$ walking to its neighbors by means of Eq. 7, and the results are shown in Fig. 3(d). From Fig. 3(d), we can see that the transition probability $p_{ij}$ is smallest, although the distance between them is nearest, whereas this probability is largest according to Eq. 20. Even if the data point $\boldsymbol{X}_i$ takes a step to $\boldsymbol{X}_j$ wrongly, and becomes closer to the point $\boldsymbol{X}_j$, its transition probability $p_{ij}$ is still smallest by means of Eq. 7, although it is bigger than its initial value, as is shown in Fig. 3(e).

# 6 Experiments and discussions

To evaluate these two clustering algorithms, we choose five datasets from UCI repository [15], which are Soybean, Iris, Wine, Ionosphere and Breast cancer Wisconsin datasets, and complete all experiments on them. In this section, firstly these datasets are introduced briefly, and then the effects of several im-portant parameters are discussed. Finally, experimental results of algorithms are illustrated.

## 6.1 Experiment setup

The original data points in above datasets all are scattered in high dimensional spaces spanned by their features, where the description of all test datasets is summarized in Table 2. As for Breast dataset, those lost features are replaced by random numbers. Finally, this algorithm is coded in Matlab 6.5.

Table 2: Description of datasets.

| Dataset | Instances | Features | Classes |
|---|---|---|---|
| Soybean | 47 | 21 | 4 |
| Iris | 150 | 4 | 3 |
| Wine | 178 | 13 | 3 |
| Ionosphere | 351 | 32 | 2 |
| Breast | 699 | 9 | 2 |

Throughout all experiments, data points in a dataset are considered as particles which can walk randomly in the whole space and whose initial positions are taken from the dataset. The similarity (or distance) measure of data points depends on the selected similarity (or distance) function $d(\cdot, \cdot)$, which satisfies the condition that the more similar data points are, the smaller the output of the function is. In experiments, the similarity (or distance) function is chosen as following:

$$d\Big(\boldsymbol{X}_i(t), \boldsymbol{X}_j(t)\Big) = exp\Big(\|\boldsymbol{X}_i(t) - \boldsymbol{X}_j(t)\|/2\sigma^2\Big), i, j = 1, 2, \cdots, N \qquad (21)$$

where the symbol $\|\cdot\|$ represents $L2$-norm. The advantages of this function are that it not only satisfies our requirements, but also it overcomes the drawbacks of Euclidean distance, for instance, when two points are too close, the output of Euclidean distance function approaches zero. However, in the modified model, transition probabilities are conversely proportional to the distances between data points according to Eq. 7. If the distance between two data points is so small that the reciprocal of the output of Euclidean distance function approaches infinite, the computation of probabilities will fail. Nevertheless, when Eq. 21 is selected as the distance function, it is more convenient to compute the transition probabilities, since its minimum is one and the reciprocals of its output are between zero and one, $1/d(\boldsymbol{X}_i(t), \boldsymbol{X}_j(t)) \in [0, 1]$. In addition, the parameter $\sigma$ in Eq. 21 and the collision-avoiding threshold $\theta$ are set one and 1.1 respectively.

Another important parameter is the interaction range $R$, which indicates the radius of neighborhood of a data point. For different datasets, it is quite difficult to preset a proper interaction range $R$ directly, because the relations of data points in different datasets are various. To simplify this problem, in experiments, we introduce a new variable $b$ to determine the interaction range $R$ indirectly. The method is as follows. At first, the initial distance matrix $D(0)_{n\times n}$ is sorted ascendingly by rows. And then the interaction range $R$ is set to the median of the $b$-th column of the distance matrix sorted, which can be expressed as below:

$$R = median\Big(sort\big(D(0)_{n\times n}\big) \times \big[0\cdots 0 \ \overset{b}{1} \ 0\cdots 0\big]^T_{1\times n}\Big) \tag{22}$$

As such, the magnitude of the interaction range $R$ may be adjusted by the variable $b$ conveniently. For example, if a big interaction range is needed, one can set a big $b$, vice versa.

## 6.2 Effects of parameters

### 6.2.1 Number of clusters vs. interaction range $R$

As is known, the interaction range $R$ or the variable $b$ controls the radius of neighborhood of a data point. For a dataset, the number of clusters depends on the interaction range $R$ partly. Generally speaking, with the increase of the interaction range $R$, the number of clusters decreases. For example, when setting a small $b$, the number of neighbors in the neighborhood of a data point $\boldsymbol{X}_i$ is small, as makes the optional transition directions reduce as well. Further, considering covers of a graph, we can find that neighborhoods of data points intersect each other slightly, so that the connected domain formed is small. Even if the data point $\boldsymbol{X}_i$ walks in space, it only observes a small area. In this case, it gathers together only with its not-too-distance data points around it. As a consequence, all data points form many small clusters at last, as is shown in Fig. 4(a).

On the other hand, if a big $b$ is selected, the data point $\boldsymbol{X}_i$ can observe a wider area because of a big interaction range $R$, at the same time that there are more neighbors in its neighborhood. Thus, the neighborhoods of data points intersect a lot and form bigger connected domains. Hence, in the end they establish several big clusters. For the same dataset, Fig. 4 exhibits the relationship between the number of clusters and the interaction range $R$. As analyzed above, when the variable $b = 10$, six clusters are formed; three clusters are established when $b = 25$. Therefore, in the case that the exact number of clusters is unknown in advance, one can adjust the interaction range $R$ or the variable $b$ to obtain different number of clusters according to practical situations.

### 6.2.2 Rate of convergence vs. the interaction range

The rate of convergence of the proposed clustering algorithms is associated with the interaction range $R$ or the variable $b$ closely. As is known, the bigger the interaction range $R$ is, the more neighbors are. In this case, according to Eq. 7, the transition probabilities of a data point $\boldsymbol{X}_i$ become smaller than those when setting a small interaction range $R$. Again, the transition distance is proportional to the transition probability, $\omega_i \propto p_{ij}$, so it drops with the decrease of transition probability as well. Thus, the rate of convergence reduces, but this also makes that data points have more chances to explore in wider space and contact with more other data points.

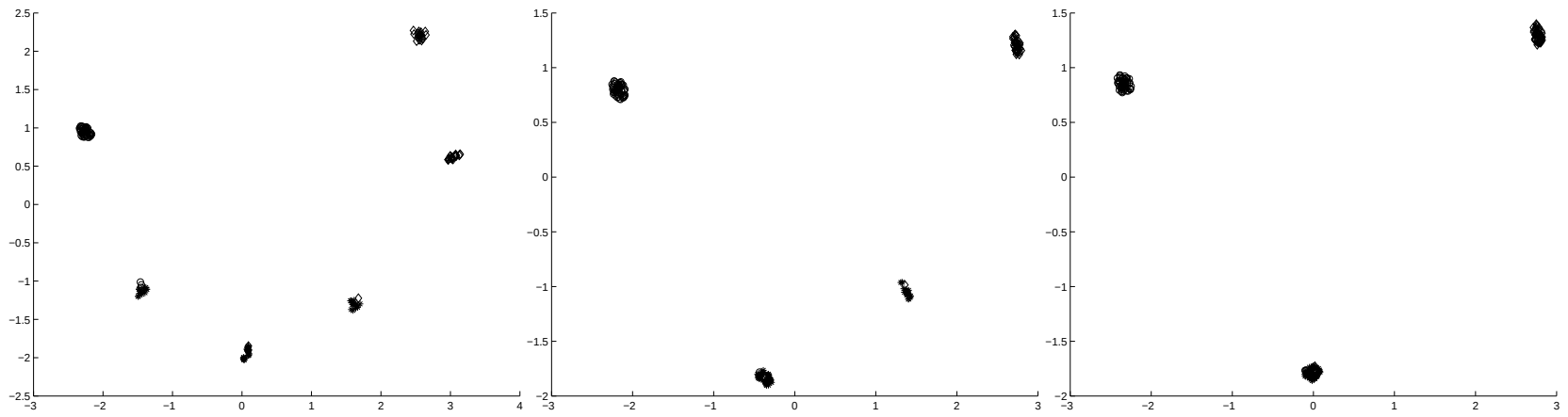

(a)$b = 10$ (b)$b = 18$ (c)$b = 25$

Figure 4: Number of clusters with the different interaction range $R$ or the variable $b$.

On the other hand, if setting a small interaction range $R$, the transition probabilities increase, at the same time that the transition distance increases too. In this case, the rate of convergence grows. However, the relationship between the rate of convergence and the interaction range $R$ needs a trade-off in order to avoid a lack of exploration or slow convergence. For the same dataset, the comparison of the rates of convergence about two clustering algorithms, RW1 and RW2, is shown in Fig. 5 with different $b$. Every dot in Fig. 5 represents the sum of transition distances of all data points $\sum_{i=1}^{n} \omega_i$ after one walking, while every dot in Fig. 5(b) is an average of results running RW2 ten times.

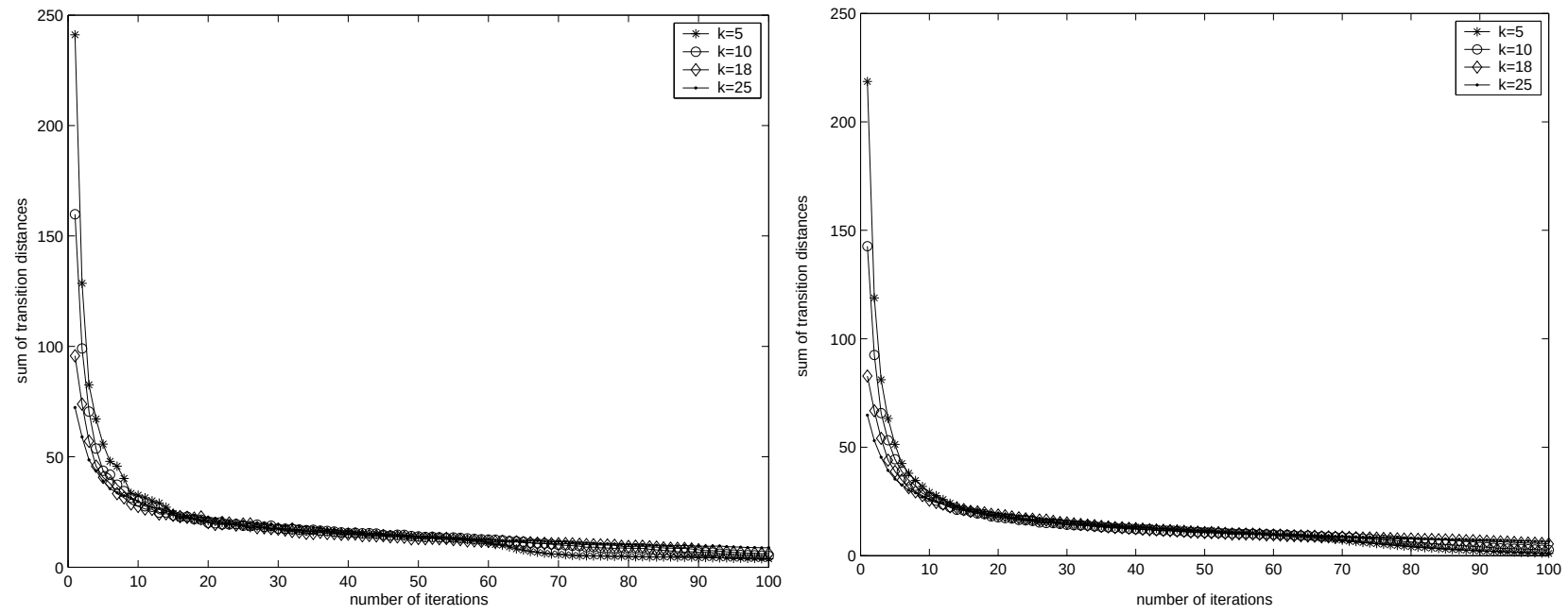

(a)Algorithm RW1 (b)Algorithm RW2

Figure 5: Comparison of the rate of convergence of two clustering algorithms.

From Fig. 5, we can see that after first walking the sum of transition distances, when $b = 5$, is much larger than the sum when $b = 25$, which means that the smaller the interaction range $R$ is, the bigger the sum of transition distances of data points is. Meanwhile, the rate of convergence decreases with the increase of $b$, but the differences are slight. Besides, the sum of transition distances of Algorithm RW2 is smaller than that of Algorithm RW1 with the same $b$, after first walking, and with the same number of iterations, Algorithm RW2 converges to a smaller value, as means in part that the exploring capacity of Algorithm RW2 is better than that of Algorithm RW1.

### 6.3 Experimental results

We have applied the two algorithms, RW1 and RW2, to above-mentioned five datasets from UCI repository. For each dataset, RW1 and RW2 are run several times to get the results at different $b$. For a same dataset, as is known, the number of clusters decreases with the increase of $b$. With a small $b$, it is possible that the number of clusters is larger than the preset number of clusters in the dataset, after the algorithm is end. So a merging-subroutine is called to merge unwanted clusters, which works in this way. At first, the cluster with the fewest data points is identified, and then is merged to the cluster whose distance between their centroids is smallest. This subroutine is repeated till the number of clusters is equal to the preset number. Finally, the results obtained by the algorithm are represented by the clustering accuracy, which is defined as below:

**Definition 1** *$cluster_i$ is the label which is assigned to a data point $\boldsymbol{X}_i$ in a dataset by the algorithm, and $c_i$ is the actual label of the data point $\boldsymbol{X}_i$ in the dataset. So the clustering accuracy is [10]:*

$$\begin{aligned} accuracy &= \frac{\sum_{i=1}^{N} \lambda\Big(map(cluster_i), c_i\Big)}{N} \\ \lambda(map(cluster_i), c_i) &= \begin{cases} 1 & \text{if } map(cluster_i) = c_i \\ 0 & otherwise \end{cases} \end{aligned} \tag{23}$$

*where the mapping function $map(\cdot)$ maps the label got by the algorithm to the actual label.*

Fig. 6(a)-(e) demonstrate the results achieved by Algorithm RW1 and RW2 on the five datasets respectively, in which every dot represents a clustering accuracy. Since Algorithm RW2 is with uncertainty, for each dataset Algorithm RW2 is run twenty times with the same $b$, and the mean and variance of those results are drawn in each figure using error bars. In addition, for each $b$, the maximum in twenty results also appears in the figure.

As is shown in Fig. 6(a)-(e), for the same dataset, the two algorithms get similar results at a same $b$, but the maximum obtained by Algorithm RW2 is much better than that of Algorithm RW1. As mentioned above, the very randomness of Algorithm RW2 may be responsible for getting better results, as also means data points have explored in wider areas.

We compare our results to those obtained by other clustering algorithms, for example, Kmeans[16], PCA-Kmeans[16] and LDA-Km[16]. The comparison is summarized in Table 3.

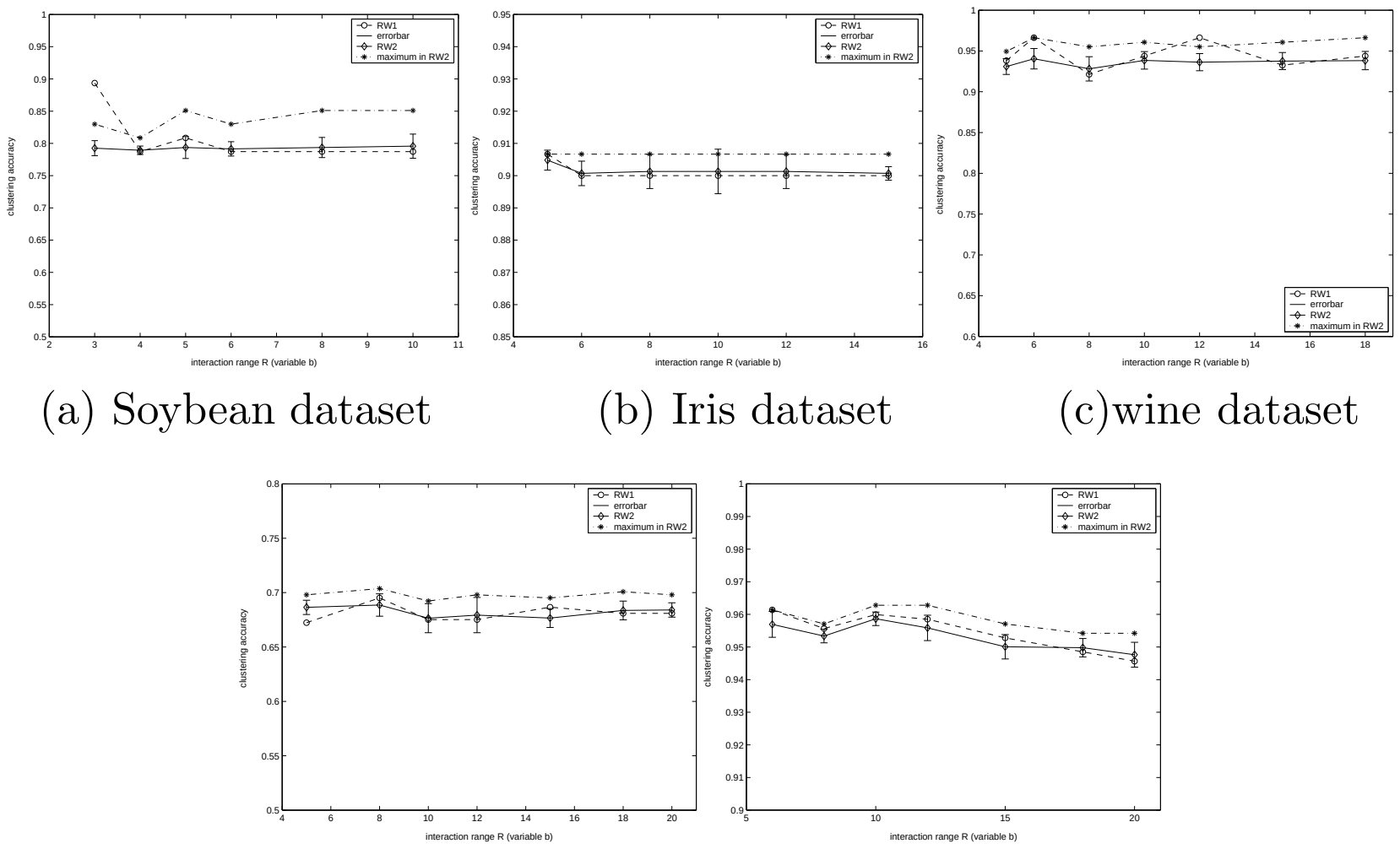

(a) Soybean dataset (b) Iris dataset (c)wine dataset

(d) Ionosphere dataset (e)breast dataset

Figure 6: Comparison of clustering accuracies of two proposed algorithms.

Table 3: Comparison of clustering accuracies of algorithm.

| Algorithm | Soybean | Iris | Wine | Ionosphere | Breast |
|---|---|---|---|---|---|
| RW1 | 89.36% | 90.67% | 96.63% | 69.52% | 96.14% |
| RW2 | 79.57±1.88 | 90.48±0.31 | 94.05±1.25 | 68.86±1.03% | 95.87±0.21% |
| MAX in RW2 | 85.11 | 90.67 | 96.63 | 70.37 | 96.28 |
| Kmeans | 68.1% | 89.3% | 70.2% | 71% | – |
| PCA-Kmeans | 72.3% | 88.7% | 70.2% | 71% | – |
| LDA-Km | 76.6% | 98% | 82.6% | 71.2% | – |

## 7 Conclusion

We have introduced a modified model of random walk, and developed two clustering algorithms based on it: (a) the deterministic random walk based clustering algorithm (RW1) and (b) the nondeterministic random walk based clustering algorithm (RW2). In those algorithms, data points in a dataset are considered as particles which can move randomly in the whole space. Initially, the sum of transition distances of data points is large, while the sum approaches a stable value when the clusters are formed gradually. If the sum is less than a preset threshold $\varepsilon$, the algorithm exits.

The modified model of random walk provides a heuristic for clustering data points. As a whole, data points tend to approach those data points with large degrees and near distances in terms of the modified model. At last, data points belonging to the same class are close to each other, and form tight clusters, while the different clusters are away from one another. If the number of clusters is unknown exactly in advance, one can adjust the interaction range $R$ or the variable $b$ to control the number of clusters, which decreases with the increases of the interaction range $R$ or the variable $b$. For the same interaction range $R$, the rates of convergence of two algorithms are fast, and Algorithm RW2 seems more exploratory.

According to Theorem 2, applying the modified model to a system with $N$ particles, some particles are close to each other, while others are away from one another. In the end, several separating clusters are formed naturally. Further, we evaluate the clustering algorithms on five real datasets, experimental results are consistent with the conclusion of Theorem 2, and data points in datasets are clustered reasonably and efficiently. In conclusion, the proposed algorithms can detect clusters with arbitrary shape, size and density.

## Acknowledgments

The authors thank Dr. Jia-qian CHEN for helpful discussion. This work is supported in part by the National Natural Science Foundation of China (No. 60405012, No. 60675055).